\documentclass{article}
\usepackage{nips14submit_e,times}

\usepackage{graphicx}
\usepackage{subfigure}
\usepackage{algorithm}
\usepackage{algorithmic}
\usepackage{times}
\usepackage{latexsym}
\usepackage{multirow}
\usepackage{url}
\usepackage{subfigure}
\usepackage{array}
\usepackage{sparklines}
\usepackage{amsmath,amsfonts,amssymb,amsthm}
\usepackage{paralist}
\usepackage{color}

\title{Learning Efficient Representations for Reinforcement Learning}
\author{Yanping Huang \\
University Of Washington\\
huangyp@cs.washington.edu}

\nipsfinalcopy 

\newcommand{\setN}{\mathcal{N}}
\newcommand{\setT}{\mathcal{T}}
\newcommand{\setS}{\mathcal{S}}
\newcommand{\setA}{\mathcal{A}}
\newcommand{\setD}{\mathcal{D}}
\newcommand{\setP}{\mathcal{P}}
\newcommand{\setR}{\mathcal{R}}
\newcommand{\E}{\mathbb{E}}
\newcommand{\D}{\mathbb{D}}
\newcommand{\R}{\mathbb{E}}
\newcommand{\Gaussian}{\mathbb{N}}

\newcommand{\norm}[1]{\left\lVert#1\right\rVert}
\newcommand{\size}[1]{\left|#1\right|}

\newcommand\argmin{\mathop{\mbox{{\rm argmin}}}\limits}
\newcommand\argmax{\mathop{\mbox{{\rm argmax}}}\limits}

\newcommand{\bigO}O

\begin{document}
\maketitle
\begin{abstract}
Markov decision processes (MDPs) are a well studied framework for solving sequential decision making problems under uncertainty. Exact methods for solving MDPs based on dynamic programming such as policy iteration and value iteration are effective on small problems. In problems with a large discrete state space or with continuous state spaces, a compact representation is essential for providing an efficient approximation solutions to MDPs.  Commonly used approximation algorithms involving constructing basis functions for projecting the value function onto a low dimensional subspace, and building a factored or hierarchical graphical model to decompose the transition and reward functions.   However, hand-coding a good compact representation for a given reinforcement learning  (RL) task can be quite difficult and time consuming. Recent approaches have attempted to automatically discover efficient representations for RL.

In this thesis proposal, we discuss the problems of automatically constructing structured kernel for kernel based RL, a popular approach to learning non-parametric approximations for value function. We explore a space of kernel structures which are built compositionally from base kernels using a context-free grammar. We examine a greedy algorithm for searching over the structure space. To demonstrate how the learned structure can represent and approximate the original RL problem in terms of compactness and efficiency, we plan to evaluate our method on a synthetic problem and compare it to other RL baselines. 
\end{abstract}

\section{Introduction}

This report considers sequential decision making problems where decisions can have both immediate and long-term effects. Each decision results in some immediate reward or benefit, but also affects the environment in which further decisions are to be made and thus affects the expected reward incurred in the future. The objective of the decision maker is to choose decision making policies optimally, that is, to maximize some long-term cumulative measurement of rewards.  Such objective is challenging mainly because of the tradeoff between upfront and future rewards. Markov decision processes~\cite{MDP1994,Mausam2012} (MDPs) provides a mathematical formalization for this tradeoff. 

\subsection{Markov Decision Process}
A MDP is mathematically defined in terms of a tuple ($\setS, \setA, \setP, \setR$), where
\begin{itemize}
\item $\setS$ is the finite set of all possible states that describes the context of the environment, also called the \textit{state space};
\item $\setA$ is the finite set of all actions the decision making agent can take;
\item $\setP: \setS \ \times \ \setA \ \times \  \setS \ \to \ [0,1]$ is a transition function, a mapping specifying the probability $P_{s,s'}^a$ of going to state $s'$  when performing action $a$ in state $s$. An essential assumption made in the MDP is that the dynamics of state evolution is \textit{Markovian}, meaning that the distribution of the next states is conditionally independent of the past, given the current state. 
\item $\setR: \setS \ \times \ \setA \ \times \setS \ \to \ \mathbb{R}$ is a reward function.  $R_{s,s'}^a$ describes a \textit{finite} payoff or reward obtained when the agent goes from state $s$ to state $s'$ as a result of executing action $a$. The reward can be either positive or negative, representing an utility or a cost, respectively. 
\end{itemize}

The optimality objective is to find a way or a \textit{policy} to maximize some measure of the long turn reward received.  A (stationary) policy $\pi:\ \setS \to \setA$ is  a mapping from states to action, which specifies an action to be taken for each state. The choice of action is independent of the time, depends only on the state. Given a policy, we can define a \textit{value} function $V_\pi(s)$ on the state space, which is the expected long run value an agent could expect to receive by choosing the action dedicated by the policy. A policy $\pi_1$ is said to dominate another policy $\pi_2$ if , $V_{\pi_1}(s) \le V_{\pi_2}(s)$  for any state $s \in \setS$, and $\exists s_1 \in \setS$ such that $V_{\pi_1}(s_1) < V_{\pi_2}(s_1)$. A fundamental theorem~\cite{Bellman1957} in MDP stated that there exists a stationary policy $\pi^*$, called the optimal policy, that dominates or has equal value to all other policies.  The existence of such an optimal policy relies on the assumption that the expected long term reward, which is the objective function in the MDP, accumulates additively over time. That is to say, at each state, the optimal policy ranks the actions based on the sum of the expected rewards of the current time step and the optimal expected rewards of all subsequent steps. 

To ensure the value function is well defined, one can limit the MDP to a finite number of time steps. In this case, the summation over rewards incurred in subsequent time steps terminates after a finite number of terms $N$, called the \textit{horizon}, and the corresponding MDP is called a \textit{finite horizon} MDP. The value of a policy $\pi$, starting from an initial state $s_0$, is
\begin{align}
 V_{\pi}^N(s) = \E [ R(s_N) + \sum_{k=0}^{N-1} R(s_k, \pi(s_k), s_{k+1} )\ |\ s_0 = s ]
\end{align}

where $R(s_N)$ is a terminal reward for ending up with the final state $s_N$, and the expectation is taken with respect to the probability distribution of the Markov Chain $\{s_0, s_1, \ldots, s_N\}$ starting at the initial state $s$, with transition probability matrix $P_{s_k,s_{k+1}}^{\pi(s_k)}$. The optimal value function and the optimal policy is denoted by $V^{*N}(s)$ and $\pi^*(s)$, respectively; that is,
\begin{align}
V^{*N}(s) &= \max_{\pi} v_{\pi}^N(s) \\ 
\pi^*(s) &= \argmax_{\pi} v_{\pi}^N(s)
\end{align}

Despite the simple mathematical properties of the finite horizon MDPs, in many tasks, the reward is accumulated over an infinite (or indefinite) sequence of time steps. We refer this kind of tasks as the \textit{infinite horizon} problems. There are three principal classes of infinite horizon problems. 
\begin{enumerate}[(a)]

\item \textbf{Discounted problems}. Here we introduce a discount factor $\gamma$ with $0 \le \gamma < 1$. The reward incurred at the $t$th transition is \textit{discounted} by a factor $\gamma^t$. Then the value function over an infinite number of time steps is given by
\begin{align}
 V_{\pi}(s) = \E [\sum_{k=0}^{\infty} \gamma^k R(s_k, \pi(s_k), s_{k+1} )\ |\ s_0 = s ]
\label{eq:value_func}
\end{align}
In our assumption, the one step reward $R_{ss'}^a$ is bounded from above by some constant, say, $M$. Therefore, $v_{\pi}(s) \le \sum_{t=0}^{\infty}\gamma^t M = \frac{M}{1 - \gamma}$, the infinite sum of decreasing geometric progression is finite for all policies $\pi$ in all situations.

\item \textbf{Stochastic Shortest Path Problems}. Here $\gamma = 1$ but we assume that there exists some additional termination state. Once the Markov chain reaches the termination state it remains there without any further rewards. The rewards (costs) associated with other states are negatively.   In addition, the Markov chain is assumed to be such that termination is inevitable within finite number of steps, at least under an optimal policy. Thus, the problem is in effect a finite horizon one, but the length of horizon may be random. It can be shown that any discounted problems can be converted to a stochastic shortest path problem.
    
\item \textbf{Average reward problems}. Without the discount factor, the sum over an infinite sequence of rewards may be infinite, however, it turns out that in many problems the average reward per time step, given by
\begin{align}
  \tilde{V}_{\pi}^N \lim_{N \to \infty}\frac{1}{N} V_{\pi}^N (s) 
\end{align} 
where $V_{\pi}^N (s)$ is the $N$-horizon value function of policy $\pi$ starting at state $s$, is well defined as a limit and is finite. \end{enumerate}

The optimal value function $V^*(s)$ can be shown to satisfy the well known \textit{Bellman equation}
\begin{align}
  V^*(s) = \max_{a\in \setA} \E[ R(s,a, s') + \gamma V^*(s')].
  \label{eq:bellman_intro}
\end{align} 

\subsection{Representations of MDPs}
Exact solutions to MDP, such as value iteration~\cite{NDP1996}, policy iteration~\cite{Howard1960}, and linear programming~\cite{DeFarias2003}, involve a \textit{lookup table representation} of the value function, in the sense that the whole vector $V(s)$ is kept in memory for each state $s$. The complexity of these algorithms are at least polynomial~\cite{Papadimitriou1987} in the size of the state space  $\size{\setS}$ as well as the size of action space $\size{\setA}$. However, the order of the polynomials is large enough that those exact algorithms are not efficient in practice. The computation requirements of large scale MDP are still overwhelming. In such problems a sub-optimal approximation solution using  \textit{compact representation} of MDPs needed to be used. compact representations for approximately solving MDPs.  Widely used  compact representations include 
\begin{itemize}
\item Construct a low dimensional vector space representation of the value function by building a set of linear basis functions~\cite{Bertsekas2000}. 
\item Kernel (instance) based methods ~\cite{Ormoneit1999} that represent the value function as a convex combination of observed values in the simulation samples.
\item Factored MDPs~\cite{Boutilier1999} construct a representation of the state space using a vector of state variables, and represent the transition models between state variables using a dynamic Bayesian network.
\item Hierarchical representations~\cite{Dayan1993,Dietterich2000} of MDPs exploit the task structure, where the actions are temporally extended. 
\item Symbolic representation of MDPs express the state space as  binary decision diagrams(BDD) and algebraic decision diagrams(ADD)~\cite{Hoey1999}.
\end{itemize}
However, finding a good compact representations for a given reinforcement learning  (RL) task requires carefully hand-coding by a human designer, which can be quite difficult and time consuming. We  further review recent developments in  automatic discovery of efficient representations in MDPs. We elaborate the problems of automatically constructing structured kernel for kernel based RL, a popular approach to learning non-parametric approximations for value function. We provide algorithms for exploring a space of kernel structures which are built compositionally from base kernels using a context-free grammar, and greedy algorithms for searching over the structure space.

\newpage
\section{Solutions for a Lookup Table Representation}
\label{sec:dp}
In this section, we review basic solutions to MDP with a lookup table representation of value function.  

There are two fundamental classes of exact solution methods to MDPs. The first approach is based on iterative algorithms that use dynamic programming, whereas the second approach formulates an MDP as a linear program. These exact solutions require a perfect knowledge of the explicit  models of the reward structure and  transition probabilities of the system, which many not be available. Simulation methods based on Monte Carlo simulations, instead requires only sample transitions   $(s_t,a_t,r_t,s_{t+1})$ of the system. 

The iterative algorithms typically employs the Bellman equation~\ref{eq:bellman_intro} to recursively relating the value of the current state to values of adjacent states. The form of Bellman equation motivates the introduction of two essential operators, also known as Bellman backup or dynamic programming backup operators in literature,  that provide a convenient shorthand notation in expressions.

For any vector $V = (V(1), \ldots, V(\size{S}))$, we consider the vector $T V$  obtained by applying one iteration of right hand side of Bellman equation:
\begin{align}
  (T V)(s) = \max_{a\in \setA} \sum_{s'\in \setS} p_{ss'}^{a} (R(s,a,s') + \gamma V(s'))
\end{align}
and similarly, for any vector $V$ and any stationary policy $\pi$, we consider the vector $T_{\pi} V$ with components
\begin{align}
  (T_{\pi} V)(s) = \sum_{s'\in \setS} p_{ss'}^{\pi(s)} (R(s,\pi(s),s') + \gamma V(s')) 
\end{align}
Given a stationary policy $\pi$, we define the $\size{\setS} \times \size{\setS}$ matrix $P_{\pi}$ whose $(i,j)$ entry is $p_{i,j}^{\pi(i)}$. Then we can re-write $T_{\pi} V$ in matrix form as
\begin{align}
  T_{\pi} V = R_{\pi} + \gamma P_{\pi} V
\end{align}
where 
\begin{align}
  R_{\pi}(s) = \sum_{s'\in \setS} p_{ss'}^{\pi(s)} R(s, \pi(s), s')
\end{align}

We denote $T^k$ and $T_{\pi}^k$ as the operator obtained by applying the mapping $T$ and $T_{\pi}$ with themselves $k$ times, respectively. It can be shown~\cite{Bertsekas2000} that the following properties hold for $T_{\pi}$ and $T$. 

\begin{enumerate}[(a)]
\item The optimal value vector $V^*$ is the only solution to the equation $V = T\ V$.
\item We have $\lim_{k\to \infty} T^k V = V^*$. for every vector $V$
\item A stationary policy is optimal if and only if $T_{\pi} V^* = T V^*$. 
\item For every vector  $V$, we have   $\lim_{k\to \infty} T_{\pi}^k V = V_{\pi}$. And $V_{\pi}$ is the only solution of the equation $V = T_{\pi}\ V$
\item The operator $T$ is a contraction mapping with respect to a weighted maximum norm. That is, there exists a vector $\rho$ of size $\size{\setS}$ and a positive scalar $\beta < 1$ such that
\begin{align}
  \norm{TV - TV'}_{\rho} \le \beta \norm{V - V'}_{\rho}
\end{align}
for all vectors $V$ and $V'$, and the weighted maximum norm is $\norm{V}_{\rho} = \max_{s\in \setS} \frac{\size{V(s)}}{\rho(s)}$ 
\end{enumerate}

\subsection{Value Iteration}
A principal method, called  value iteration, for calculating the optimal value $V^*$ is to generate a sequence $T^k V$ starting from some vector $V$ as $\lim_{k\to \infty} T^k V = V^*$. The value functions so computed are guaranteed to converge in the limit to the optimal value function.  In the stochastic shortest path  and average reward problems some additional assumptions for convergence are needed. 
\begin{itemize}
\item \textit{Finite (N) horizon  problem}: the algorithm always converge in $N$ steps. 
\item \textit{Infinite horizon problems with discount rewards}: the algorithm always converges to the unique optimal solution.
\item \textit{Stochastic shortest path problem}: the algorithm converges if there is a policy with positive probability of termination after at most finite time steps, regardless the initial state. 
\item \textit{Average Reward problems}: the algorithm converges if every state can be reached from every other state in finite time step with positive probability for some policy. 
\end{itemize}

\begin{algorithm}
  \caption{Value Iteration}
  \label{alg:valueIteration}
  \begin{algorithmic}[1]
    \STATE Initial $V_0$ arbitrarily for each state and $t = 0$
    \REPEAT
        \STATE  Compute $V_{t} = T V_{t-1}$
        \STATE  Compute Residual $e_t = \norm{V_t - V_{t-1}}_{max}$
        \STATE  $t = t\ +\ 1$
    \UNTIL $e_t < \epsilon$
    \RETURN Greedy policy $\pi(s) = \argmax_{a\in \setA} \sum_{s'\in \setS} P_{ss'}^{a}[R(s,a,s') + \gamma V_t(s')]$
  \end{algorithmic}
\end{algorithm}
A commonly used stopping rule is to set $\epsilon = \epsilon' \frac{1 - \gamma}{ 2 \gamma}$, which ensures the resulting value function is within $\frac{\epsilon'}{2}$ of the optimal value function, and the resulting policy is $\epsilon'$-optimal~\cite{Williams1993}. 

The running time for each iteration in algorithm\ref{alg:valueIteration} is $\bigO(\size{\setA}\size{\setS}^2)$. The number of iterations until convergence it shown~\cite{Littman1995} to be polynomial in the size of the state space  $\size{\setS}$ as well as the size of action space $\size{\setA}$, which in turn makes value iteration polynomial in time.   However, the order of the polynomials is nontrivial, thus in practice value iteration is usually inefficient. 

\subsection{Policy Iteration}
Another widely used iterative algorithm is known as policy iteration~\cite{Howard1960}. At each iteration, the decision maker first carries out a \textit{policy evaluation} phase, in which the value function associated with the current policy is computed, and a \textit{policy improvement} phase, in which a greedy attempt is made to improve the current policy. 

The basic policy iteration algorithm is described in algorithm~\ref{alg:policyIteration},
\begin{algorithm}
  \caption{Policy Iteration}
  \label{alg:policyIteration}
  \begin{algorithmic}[1]
    \STATE Let $\pi_0$ be some random initial policy  and $t = 0$
    \REPEAT 
    \STATE Policy Evaluation: compute $V_{\pi_t}$ in equation~\ref{eq:policyEvaluation}.
    \STATE Policy Improvement:  $\pi_{t+1}(s) = \argmax_{a\in \setA} \sum_{s'} P_{ss'}^a (R_{ss'}^a + \gamma V_{\pi_t}(s'))$, for all $s \in \setS$
        \STATE  $t = t\ +\ 1$
    \UNTIL $\pi_{t+1}(s) = \pi_t(s)$, for all  $s \in \setS$
  \end{algorithmic}
\end{algorithm}
where policy evaluation step involves solving a system of $\setS$ equations with $\setS$ unknowns. Let $\rho$ be the invariant distribution of a Markov chain $P_{\pi}$, and let $\setN$ be the set of non-terminal states and $\setT = \setS - \setN$ be the set of zero reward termination states in stochastic shortest path problems.  

\begin{align}
     \begin{array}{rll}
    V_{\pi}(\setN) &=  (I - P_{\pi}(\setN, \setN))^{-1} (R_{\pi}(\setN) + P_{\pi}(\setN, \setT) R_{\pi}(\setT) )  & \mbox{Stochastic Shortest Path}\\
     V_{\pi} & =  (I - \gamma P_{\pi})^{-1} R_{\pi} & \mbox{Discounted Reward} \\
\tilde{V}_{\pi} &= (1 - P_{\pi})^{-1} (R_{\pi} - \rho) & \mbox{Average Reward}
      \end{array}
   \label{eq:policyEvaluation}
\end{align}

For each iteration, policy evaluation phase can be performed in $\bigO(\size{\setS}^3)$ arithmetic operations and policy improvement in $\bigO(\size{\setA}\size{\setS}^2)$ operations. When the number of states is large, it's usually preferable to carry out the policy evaluation phase by using iterative methods such as value iteration. It can be shown that the policy iteration algorithm generates an improving sequence of policies and terminates with an optimal policy. There is no theoretical guarantees for the number of iterations required, yet policy iteration has been listed as one of the preferred solution method for MDP.  

\subsection{Linear Programming}
A third approach to solve MDPs exactly is based on linear programming~\cite{DeFarias2003}. The primal linear program involves
\begin{align}
  \begin{array}{rl}
       \mbox{Variables: } & V(s), \quad \forall s \in \setS \\
       \mbox{Minimize: } & \sum \rho(s) V{s} \\
       \mbox{Subject to: } & V(s) \ge \sum_{s'} P_{ss'}^a (R_{ss'}^a + \gamma V_{\pi_t}(s'))\quad \forall s \in \setS, \forall a \in \setA
      \end{array}
\end{align}
where $\rho$ is known as the state relevance weight vector whose elements are all positive. There are $\size{\setA}\size{\setS}$ constraints and $\size{\setS}$ variables, one constrainst for each state $s$ and action $a$. Thus, MDPs can be solve in  polynomial time.  A drawback of this algorithm is that it is typically slower than those iterative dynamic programming methods.

\subsection{Temporal Difference Learning}
In this subsection, we discuss an implementation of the Monte Carlo algorithm that incrementally updates the value function $V(s)$ after each transition. We first express the value function as 
\begin{align}
  V_{\pi}(s_t) &= \E [\sum_{m=0}^{\infty} \gamma^m g(s_{t+m}, s_{t+m+1})] \nonumber \\
 &= \E[g(s_t,s_{t+1}) + \gamma V_{\pi}(s_{t+1}) ]
\end{align}
The Robbins-Monro stochastic approximation method for solving the above expectation equation takes the form
\begin{align}
  \hat{V}(s_t)  &= (1 - \alpha_t) \hat{V}(s_t) + \alpha_t (g(s_t, s_{t+1}) + \gamma \hat{V}(s_{t+1}) - \hat{V}(s_t) ) \nonumber \\
& =  (1 - \alpha_t) \hat{V}(s) + \alpha_t d_t
\end{align}
where $\alpha_t \in (0,1)$ is the learning rate and $d_t = g(s_t, s_{t+1}) + \gamma \hat{V}(s_{t+1}) - \hat{V}(s_t)$ is called the temporal difference (TD)~\cite{Sutton1988}, representing the difference between an estimate $g(s_t, \pi(s_t), s_{t+1}) + \gamma \hat{V}(s_{t+1}) $ of the value function based on the one-step ahead simulated outcome of the current time step, and the current estimate $\hat{V}(s_t)$. Alternatively, we might fix a non-negative integer $L$ and take into accounts the $L+1$-step ahead simulated outcome,
\begin{align}
   V_{\pi}(s_t) &= \E[\sum_{m=0}^L \gamma^m g(s_{t+m}, s_{t+m+1}) + V_{\pi}(s_{t+L+1})]
\label{eq:LstepsBellmanEq}
\end{align}
We cannot assume one $L$ better than another in the absence of any special knowledge. For the sake of generality, we may combine a weighted average of $L$-step Bellman equation~\ref{eq:LstepsBellmanEq} over all possible $L$. We introduce a constant $\lambda < 1$, multiply Eq.\ref{eq:LstepsBellmanEq} by $(1-\lambda)\lambda^L$, and sum over all non-negative $L$. We then have,
\begin{align}
   V_{\pi}(s_t) &= (1 - \lambda) \E [ \sum_{L = 0}^{\infty} \lambda^L (  \sum_{m=0}^L \gamma^m g(s_{t+m}, s_{t+m+1}) + V_{\pi}(s_{t+L+1}) )
] \nonumber \\
&= \E[(1-\lambda)\sum_{m=0}^{\infty} g(s_{t+m}, s_{t+m+1})\sum_{L=m}^{\infty} \lambda^m + \sum_{L=0}^{\infty} (\lambda^{L} - \lambda^{L+1}) V_{\pi}(s_{t+L+1}) ] \nonumber \\
 &= \E[ \sum_{m=0}^{\infty} \lambda^m \gamma^m d_{m+t} ] + V_{\pi}(s_t)
\end{align}
The resulting Robbins-Monro stochastic approximation method is then
\begin{align}
  \hat{V}(s_t) = (1 - \alpha_t) \hat{V}(s_t) + \alpha_t \sum_{m=t}^{\infty} (\lambda\gamma)^{m-t} d_m
\label{eq:TD_lambda}
\end{align}
The above equation provides a family of algorithms, one for each $\lambda$, and is known as TD($\lambda$). The choice of $\lambda$ reflects a trade-off between bias and variance in the Monte Carlo based approximation.  The general conclusion from ~\cite{SinghD1996} shows that intermediate values of $\lambda$ seem to work best in practise.  Sutton~\cite{Sutton1988} has shown that under TD($0$), the temporal difference algorithm converges to the true value function $V_{\pi}$. Dayan~\cite{Dayan1992} extended this result to the case of general $\lambda$.

A temporal difference based method for learning action values called Q-learning was introduced by Waktins~\cite{watkins92}. Q-learning updates directly estimates of the Q-factors associated with an optimal policy, thereby avoiding the multiple policy evaluation phases of policy iteration. The following learning rule for learning the action value function $Q(s,a)$ is used:
\begin{align}
  Q_{t+1}(s,a) = (1 - \alpha_t) Q_t(s,a) + \alpha_t (g(s,a,s') + \gamma \max_{a'\in \setA} Q_t(s',a'))
\label{eq:qlearning}
\end{align}
where $s'$ and $g(s,a,s')$ are generated from the pair $(s,a)$ by simulation, according to the transition probability matrix $P_{ss'}^a$. Q-learning is sometimes referred to as an \textit{off-policy} learning algorithm since it estimates the optimal action value function $Q(s,a)$ while simulation the MDP using any policy. During simulation, a sequence of states is generated with the greedy actions provided by the current available Q-factors. It's possible that certain profitable actions are never explored. In practice, variants of Q-learning algorithms with parameters control the degree of exploration are introduced to ensure sufficient exploration during simulations.
\newpage
\section{Compact Representation of  Markov Decision Processes}
The solutions described in previous section require a lookup table representations of the value function $V(s)$ with size $\size{\setS}$.
In environments with large discrete state space is large or even with continuous state spaces, the time complexity of the MDP solution algorithms makes them inefficient in practise. In this section, we review a variety of compact representations for approximately solving MDPs, including low dimensional vector space representations by constructing linear basis functions~\cite{Bertsekas2000}, instance based representations of value function using kernels in Hilbert space~\cite{Ormoneit1999}, factored representation~\cite{Guestrin2003},  hierarchical representations~\cite{Dayan1993,Dietterich2000}, and symbolic representations such as binary decision diagrams(BDD) and algebraic decision diagrams(ADD)~\cite{Hoey1999}. All these approaches depend crucially on a choice of low dimensional compact representation of a MDP, and assume these are carefully provided by the human designer. The focus of this section is on approximation, rather than automatic representation discovery.

\subsection{Linear Value Function Approximation}
In this subsection, we consider the policy evaluation phase for a single stationary policy $\pi$. Thus we suppress in our notation for the value functions the dependence on $\pi$. We approximate the value function $V(s)$ with a linear architecture:
\begin{align}
  \hat{V}(s, w) = \phi(s)' w, \quad \forall s \in \setS
\end{align}
where $w$ is a weight vector and $\phi(i)$ is an $\size{\setD}$-dimensional feature vector associated with state $s$. That is, we represent the value function in a compact form $V \approx \hat{V} = \Phi w$, where $\Phi$ is the $\size{\setS} \times \size{\setD}$ matrix that has as rows the feature vectors $\phi(s)$, $s \in \setS$. Thus, we want to approximate the value function $V$ with the subspace $\setD$ spanned by $\size{\setD}$ basis function, each of which is in the columns of $\Phi$. The rank of matrix $\Phi$ is $\size{\setD}$.  Let $\Pi$ be the projection operator on to the linear subspace, with respect to some norm $\norm{\cdot}_{\rho}$:
\begin{align}
  \norm{V}_{\rho} = \sqrt{\sum_{s\in \setS} \rho_s V^2(s)},
\end{align}
where $\rho$ is a vector of positive components. $\Pi V$ is the unique vector in the subspace that minimizes $\norm{V - \Phi\ w}_{\rho}$. 
\begin{align}
  \Pi V &= \Phi\ w_{\Phi} \\
  w_{V} &= \argmin_{w \in \R^{\D}} \norm{V - \Phi\ w}_{\rho}^2 \label{eq:leastSqureProj}
\end{align}
By setting the gradient of Eq.~\ref{eq:leastSqureProj} to 0, we have
\begin{align}
  \Pi = \Phi (\Phi' D_{\rho} \Phi)^{-1} D_{\rho}
  \label{eq:projOperator}
\end{align}
where $D_{\rho}$ is the $\size{\setS} \times \size{\setS}$ diagonal matrix whose entries are  $\rho(s)$. Now consider the Bellman backup operator $T_{\pi}$ updating projected value functions,
\begin{align}
  \Phi\ w &= \Pi T_{\pi} (\Phi\ w) \nonumber \\ 
  \Phi\ w &= \Pi [R_{\pi} + \gamma P_{\pi} \Phi\ w ]
\end{align}
This equation is known as the projected Bellman's equation. And the solution $\phi\ w_{\Phi}$ of this equation is the approximation to value function $V_{\pi}$ in the subspace spanned by $\Phi$. $w_{\Phi}$ satisfied 
\begin{align}
[ \Phi' D_{\rho} (I - \gamma P_{\pi}) \Phi ]\ w_{\phi} &= \Phi' D_{\rho}  R_{\pi} \nonumber  \\
A w_{\phi} & = b \label{eq:projectedBellman}
\end{align}
and can be solved by matrix inversion $w = A^{-1} b$ or other iterative algorithms.  It can be shown that both mapping $T_{\pi}$ and $\Pi T_{\pi}$ are contraction~\cite{Munos2003} with respect to the weighted Euclidean norm $\norm{\cdot}_{\rho}$, where $\rho$ is the steady state probability vector of the Markov chain with transition probabilities $P_{\pi}$. Analog to value iteration, the so-called projected value iteration algorithm iteratively apply the contraction operator $\Pi T_{\pi}$, starting with some arbitrary vector $w_0$
\begin{align}
  \Phi\ w_{t+1} = \Pi T_{\pi} (\Phi\ w_t)
\label{eq:projectedValueIteration}
\end{align}
However, the projected value iteration algorithm is not practical when $\size{\setS}$ is large since $T_{\pi} (\Phi\ w_t)$ is of size $\size{\setS}$, and the steady state probabilities $\rho$ are assumbed to be known. 

Alternative way to solve equation~\ref{eq:projectedBellman} from simulation trajectories sampled from the Markov chain associated with policy $\pi$. After collecting $t$ samples we have 
\begin{align}
  \hat{A}_t &= \frac{1}{t+1} \sum_{k=0}^t \phi(s_k) (\phi(s_{k}) - \gamma \phi(s_{k+1}))'\\
\hat{b}_t &= \frac{1}{t+1} \sum_{k=0}^t \phi(s_k) R(s_k, s_{k+1})
\end{align}
Given $\hat{A}_t$ and $\hat{b}_t$, one can construct a simulation bases solution
\begin{align}
  w_t =  \hat{A}_t^{-1}  \hat{b}_t
\end{align}
This is known as the least square temporal difference (LSTD) method. 

Similar to TD($\lambda$) method, we can introduce a constant $\lambda < 1$ and define
\begin{align}
  \hat{A}_t^\lambda &= \frac{1}{t+1} \sum_{k=0}^t \phi(s_k) \sum_{m=k}^t \gamma^{m-k} \lambda^{m-k} (\phi(s_m) - \gamma \phi(s_{m+1}))'\\
\hat{b}_t^\lambda &= \frac{1}{t+1} \sum_{k=0}^t \phi(s_k) \sum_{m=k}^t \gamma^{m-k}\lambda^{m-k} R(s_m, s_{m+1})
\end{align}
the corresponding matrix inversion solution $w_t = ( \hat{A}_t^\lambda )^{-1} \hat{b}_t^\lambda$ is called the LSTD($\lambda$) method.

\subsection{Factored Markov Decision Processes}
When some structure knowledge about the state space is known, one can construct a \textit{factored MDP} representation of the state space using a vector of state variables, and represent the transition models between state variables using a dynamic Bayesian network. In this way, the value function can be approximated by a linear combination of basis functions, where each basis function involves only a small  subset of the state variables. In particular, Guestrin et al~\cite{Guestrin2003} proposed an algorithm that generalize exact linear programming using basis functions $\Phi$. 

\begin{align}
  \begin{array}{rl}
       \mbox{Variables: } & w_1, \ldots, w_{\size{\setD}} \\
       \mbox{Minimize: } & \sum_s \rho(s) \sum_{i}w_i \phi_i{s} \\
       \mbox{Subject to: } & \sum_i w_i \phi_i(s) \ge \sum_{s'} P_{ss'}^a (R_{ss'}^a + \gamma \sum_i w_i \phi_i(s') )\quad \forall s \in \setS, \forall a \in \setA
      \end{array}
\end{align}
where $\rho$ is known as the state relevance weight vector whose elements are all positive. The number of variables in linear program has now been reduced from $\size{\setS}$ to $\size{\setD}$, the number of basis function in sub-space $\setD$. Without a factored representation of the state space, the number of constraints remains $\size{\setS} \times \size{\setA}$. For factored MDPs, the number of constraints can be reduced exponentially by exploiting conditional independence properties in the conditional probability table of the dynamic Bayesian network.

\subsection{Kernel Based Reinforcement Learning}
In the kernel based reinforcement learning (KBRL) algorithms~\cite{Ormoneit1999,Jong2006},  value functions are approximated by a set of sample outcomes $\{s_t, a_t, r_t, s_{t+1}\}_{t=1}^{N_T}$. Specifically, KBRL approximates the outcome of an action $a$ from a given state $s$ as the convex combination of sampled outcomes of that action, weighted by a function of the distance between $s$ and sampled states. Then
the Bellman backup operator is represented by an operator $T_K$ on the samples:
\begin{align}
\hat{V}(s) &=  T_K V(s) =  \max_{a\in \setA} \hat{Q}(s,a)\\
  \hat{Q}(s,a) &= \sum_{t \in \{t: a_t = a \}} K_a(s_t,s) [r_t + \gamma V(s_{t+1}) ] 
\end{align}
where the summation is over a subset of indices $t$  where $a_t = a$, and the kernel $K_a(s_t,s)$ is normalized in the sense that for each state $s$ and action $a$, $\sum_{t \in \{t: a_t = a \}} K_a(s_t, s) = 1$. 

Kernel-based reinforcement learning has several promising properties. First, the operator $T_K$ has a unique fixed point. One can obtain an algorithm analog to value iteration to solve the MDP by iteratively applying  $T_K$. Second, the fix point of this operator converges in probability to the true value function for the Gaussian Kernel:
\begin{align}
  K_a(s_t, s) = \exp[-\frac{d^2(s_t,s)}{2\sigma^2}]
\end{align}
when the number of samples $N_T \to \infty$ and the bandwidth $\sigma \to 0$. The distance metric $d(s_t,s)$ denotes the distance function. However, the time complexity of KBRL is $N_T^2)$, which make it impractical when the sample size is large. To make it practical, Kveton~\cite{Kveton2012} employs an unsupervised learning method to cluster the simulation samples onto $k$ representative ones, and is able to compute the optimal policy in $O(n)$ time assuming $k \ll n$ a constant regardless $n$. Another advantage of the kernel based methods is the straightforward incorporation of the structure knowledge of the state space by using the structure kernel~\cite{Kveton2013}, where the kernel $K_a(s_t,s)$ can be decomposed into a product of base kernels. 

The kernel based algorithm defined above requires knowledge about the metric function of the state space. Alternatively, the Gaussian Process Temporal Difference (GPTD)~\cite{Engel2005} learning offers a Bayesian solution. Consider an episode in which a terminal state is reached at time step $T+1$, with $r_{T+1} = V(X_{T+1}) = 0$. We have a generated model for the value function at state $s_t$:
\begin{align}
  V(s_t) = r_t + \gamma r_{t+1} + \ldots + \gamma^{T-t}r_T - \epsilon_t
\end{align}
with $\epsilon_t \sim \Gaussian(0,\sigma_t^2))$.  In a matrix form, we have
\begin{align}
  Z_T r_{1:T} &= V_{1:T} + \epsilon_{1:T} \\
  r_{1:T} &= H_{T+1} V_{1:T} + \epsilon'_{1:T} 
\end{align}
where
\begin{align}
  Z_T = \left [ 
    \begin{array}{ccccc}
      1 & \gamma & \gamma^2 & \ldots & \gamma^T \\
      0 & 1  & \gamma & \ldots & \gamma^{T-1} \\
      \hdots & & & \hdots \\
      0 & 0 &  0 & \ldots & 1
    \end{array}
 \right ] 
\quad H_{T} = Z^{-1}_{T-1} = \left [
 \begin{array}{ccccc}
      1 & -\gamma & 0 & \ldots & 0 \\
      0 & 1  & -\gamma & \ldots & 0 \\
      \hdots & & & \hdots \\
      0 & 0 &  \ldots & 1 & -\gamma
    \end{array}
\right ]
\end{align}
Assuming a state-wise noise model with $\epsilon_t \sim \Gaussian(0, \sigma^2)$, we have $\epsilon'_{1:T} \sim \Gaussian(0, \sigma^2 H_TH_T^T)$. 

Since both the value prior and the noise are Gaussian, so is the posterior distribution of the value conditioned on an observed sequence of rewards $r_{1:T} = \{r_t \}_{t= 1: T}$. The joint distribution between a test point $V(s^*)$ and the observed sequence is:
\begin{align}
  \left ( \begin{array}{c} 
      Z_T\ r_{1:T} \\ V(s^*)  
   \end{array}  \right ) = \Gaussian \left[
  \left ( \begin{array}{c} 
     0 \\ 0  
   \end{array}  \right ),
    \left [ \begin{array}{cc} 
      K_T & K_T(s^*) \\
   K_T(s^*)^T & K(s^*,s^*)
   \end{array}  \right ]
\right ]
\end{align}
where $K_T$ denotes the $T \times T$ matrix of the covariances evaluated at all pairs of observed states, and $K_T(s^*)$ denotes the $T \times 1$ vector of the covariances evaluated at pairs of observed state $s_t$ and the test state $s^*$. The posterior mean and variance of the value at $s^*$  are given, respectively, by
\begin{align}
  \hat{V}(s^*) & = K_T(s^*)^T (K_T + \sigma I)^{-1}\ r_{1:T} \\
  \rm{VAR}(\hat{V}(s^*)) &=  K(s^*,s^*) - K_T(s^*)^T (K_T + \sigma I)^{-1} K_T(s^*)
\end{align}
\subsection{Hierarchical Methods}

Another approach to solving MDPs with large state spaces is to treat them as a hierarchical of task structures. In many cases, hierarchical solutions don't aim at providing an optimal value function to a  MDP problem, but focus on gaining efficiency in execution time and learning time. Hierarchical learners are commonly structured as \textit{delegation} behaviors. Feudal Q-learning~\cite{Dayan1993} involves a hierarchy of learning problems, with higher level agents being masters and lower level agents being slaves. The highest level agent receives rewards $r_t$ and states $s_t$ from the external environment. It learns a mapping from states $s_t$ to some pre-defined intermediate commands  and feeds the lower level slaves commands and corresponding rewards for taking actions  that satisfy the command. The lower level agents learns a mapping from commands and states to external actions $a_t$. However, the set of intermediate commands and their associated reinforcement functions should be established in advance of the learning. Similarly, by assuming one can identify useful subgoals and define subtasks that achieve these subgoals, 
 the MAXQ algorithms~\cite{Dietterich2000} that decompose the target MDP into a hierarchy of smaller MDPs  were proposed. Using the MAXQ decomposition, the value function of the target MDP can be expressed as an additive combination of the value functions of the smaller MDPs. To amend restriction of human designed hierarchy, Mehta et al~\cite{MehtaRTD2011} further introduced an algorithm that can automatic discover the task hierarchy, given that the dynamic Bayesian networks  associated with the action and reward models  are provided, as well as successful sample trajectories following the optimal policy.

\subsection{Symbolic Algorithms for Solving MDPs}
We briefly discussed symbolic algorithms in this subsection. 
The key idea of symbolic algorithms is to compactly represent the MDP models (value function, transition probabilities, reward functions, etc) using decision diagrams, instead of using the table lookup representation. Similar to \textit{aggregation} methods, these decision diagram representations cluster the states that share similar values. Instead of applying Bellman operator to each state, it is sufficient to update the subset of states with similar values as a whole at once, by just a single Bellman backup. This representation allows one to describe a  value function as a function of the variables describing the domain and speeds up the value iteration based algorithms. However, these symbolic algorithms assume states in the MDP be factored. That is, the state space $\setS$ is factored into a set of $d$ boolean state variables $s = \{s_1,\ldots, s_d\}$. Although any finite valued non boolean variable can be split into a number of boolean variables, it often makes the new state space using decision diagram representation larger than the original one using the lookup table representation.

\newpage
\section{Representation Learning in Markov Decision Processes}
In this section, we discuss methods for constructing compact representation of MDPs. 

\subsection{Feature Generation through Automatic Basis Construction}
The policy evaluation phase can be viewed as solving systems of linear equation of the form $Aw = b$. The Krylov space method has long been among the most successful methods currently available for efficiently solving systems of linear equations. The $k$-order Krylov subspace is the linear subspace spanned by the image of $b$ under the first $k-1$ powers of $A$, that is,
\begin{align}
  Krylov_k(A,b) = \mathrm{span}\{b, Ab, A^2b, \ldots, A^{k-1}b \}
\end{align}
For an MDP, typically we set $b = R_{\pi}$. The Krylov basis can be significantly accelerated by a computational trick called the Schultz expansion, 
\begin{align}
  (1 - A)^{-1} b = (I + A + A^2 + \ldots ) b = \prod_{k=0}^{\infty} (I + A^{2^k}) b 
\end{align}
For example, we can compute the policy evaluation phase as follows:
\begin{align}
  V_{\pi}  = ( 1 - \gamma P_{\pi})^{-1} R_{\pi} = \prod_{k=0}^{\infty} ( I + (\gamma P_{\pi})^{2^k}) R_{\pi}
\end{align}

Another way to construct basis automatically is based on the residual error in the current feature set~\cite{Parr2007}. Formally, if $\Phi_k$ is the current set of basis functions, the Bellman error basis functions (BEBFs) add $\phi_{k+1} = R + \gamma P \Phi_k w_{\Phi_k} -  \Phi_k w_{\Phi_k} $ as the next basis function.

It's been shown~\cite{Parr2008} that a basis $\Phi$ is not only useful in approximating value functions, but also induces a \textit{low-dimensional} MDP.  The induced approximate reward function $R_{\pi}^{\Phi}$ and approximate transition function  $P_{\pi}^{\Phi}$ are defined as 
\begin{align}
   R_{\pi}^{\Phi} &= (\Phi' D_{\rho} \Phi)^{-1} \Phi' D_{\rho} R_{\pi} \\
P_{\pi}^{\Phi} &= (\Phi' D_{\rho} \Phi)^{-1} \Phi' D_{\rho} P_{\pi} \Phi
\end{align}
where $R_{\pi}^{\Phi}$ is the projection of the reward function $R_{\pi}$ onto the column space of $\Phi$, with respect to $\norm{\cdot}_{\rho}$. Similarly, $P_{\pi}^{\Phi}$ is the least square solution to the system $\Phi P_{\Pi}^{\Phi} \approx P_{\Pi} \Phi$. The exact solution to this approximate MDP is the same as that given by the exact solution to the original MDP projected onto the basis $\Phi$.

Given basis constructed by Krylov space or BEBF methods with $k$ basis functions, Mahadevan~\cite{Mahadevan2012} propose the representation policy iteration algorithm, as described in Algorithm~\ref{alg:RPI}

\begin{algorithm}
  \caption{Model-based representation policy iteration}
  \label{alg:RPI}
  \begin{algorithmic}[1]
    \STATE Let $\pi_0$ be arbitrary policy and $t = 0$
    \REPEAT 
    \STATE Construct basis matrix $\Phi$
    \STATE From the MDP compute $R_{\pi_t}^{\Phi}$  and $P_{\pi_t}^{\Phi}$
    \STATE Find the solution to $(1-\gamma P_{\pi_t}^{\Phi}) w_{\Phi} = R_{\pi_t}^{\Phi}$
\STATE Project solution back to the original state space $V_{\pi_t}^{\Phi} = \Phi w_{\Phi}$.
\STATE Find the greedy policy $\pi_{t+1}$ as in the policy improvement phase
\begin{align}
  \pi_{t+1}(s) = \argmax_{a\in \setA}\sum_{s'} P_{ss'}^a (R_{ss'}^a + \gamma V_{\pi_t}^{\Phi}(s'))
\end{align}
\STATE $t = t+1$
    \UNTIL $\pi_t = \pi_{t+1}$
    \RETURN $\pi_{t+1}$
  \end{algorithmic}
\end{algorithm}

\subsection{Feature Generation through Adaptive State Aggregation}
Another basis construction algorithm~\cite{Bertsekas1989} called the \textit{adaptive state aggregation} partitions the original state space $\setS$ into a set of $m$ subsets $\setS_1,\ldots, \setS_m$, where $\cup_{i=1}^m \setS_i = \setS$ and $\setS_i \cap \setS_j = \emptyset$, for $i\neq j$. We can view state aggregation as a special form of basis matrix $\Phi$, where each column represents an indicator function for each cluster. At each iteration, the algorithm first carries out the regular value iteration to compute $V^{k+1}$, then corrects, rather than projects, $V^{k+1}$ using the basis matrix
\begin{align}
  V^{k+1} = V^{k} + \Phi\ w_{\Phi}
\end{align}
where $w_{\Phi}$  is the solution to the compact policy evaluation problem
\begin{align}
  w_{\Phi} &= (I - \gamma P_{\Pi}^{\Phi})^{-1} R_{\Pi}^{\Phi}\\
P_{\Pi}^{\Phi} &= (\Phi' D_{\rho} \Phi)^{-1} \Phi' P_{\pi} \Phi \\
R_{\Pi}^{\Phi} &= (\Phi' D_{\rho} \Phi)^{-1} \Phi'  (T (V^{k}) - V^k)
\end{align}
To create the basis $\Phi$ automatically,  Keller~\cite{Keller2006} 
proposed to use neighborhood component analysis (NCA), a supervised learning algorithm with the state $s$ as the input attributes, and the Bellman error or the temporal difference error as the supervised signal. In this way, NCA places basis function in the lower-dimensional space. The new lower dimensional features are then added as new features for the linear function approximator. 

\subsection{Structure Learning in Factored MDPs}

\begin{algorithm}
  \caption{Structure Learning Algorithm for factored MDP}
  \label{alg:sdyna}
  \begin{algorithmic}[1]
    \STATE Initialization
    \FOR{each time step $t$}
    \STATE{Given $s, \pi_{t-1}(s)$, observe $s'$ and $r$}
    \STATE{Update the factored representation of reward Fact($R_t$) and transition Fact($P_t$) functions.}
    \STATE{Learn a policy $\pi_t$ using structure value iteration or algorithms for factored MDP.}
    \ENDFOR
  \end{algorithmic}
\end{algorithm}

Factored MDPs ~\cite{Boutilier1999,Guestrin2003} compactly represent the transition and reward functions of a MDP using dynamic Bayesian networks (DBNs). Efficient algorithms based linear program were developed even when the state space is large. However, they require a complete knowledge of the transition and reward functions of the problem in advance. Structure learning algorithms \cite{Degris2006}, as sketched in Algorithm~\ref{alg:sdyna} has been proposed to learn these functions by simulation trials, where decision tree induction algorithms are used to learn a factor representation of the reward and transition functions. Given the sample transitions $\{s_t,a_t,r_t, s_{t+1}\}$ observed in a MDP system, decision tree induction algorithms learn the compact reward model with $\{s_t\}$ being example attributes and $\{r_t\}$ being example labels, and learn a conditional probabilities table representation of the transition model with $\{s_t\}$ being example attributes and $\{s_{t+1}\}$ being example labels. A $\chi^2$ test is used to detect the independence between two random variables. After a factored representation of the model is learned incrementally, the improved policy can be obtained by an incremental version of structured value iteration~\cite{Boutilier1999}. At the next iteration, the agent will follow the $\epsilon$-greedy variant of the updated policy and generate new simulation samples. The algorithm will again update its factored representation for the model. 

\subsection{Structure Discovery through Compositional Kernel Search}
Unlike the parametric linear function approximation using basis $\Phi$, Kernel-based reinforcement learning (KBRL)~\cite{Ormoneit1999, Rasmussen04} is a popular approach to learning a non-parametric representation of the value function, where the similarities between two states are captured by a kernel $K_a(s,s')$. In problems where the state space is factored and $s$ can be expressed as a set of state variables, among which there exists some conditional independencies, structured kernels~\cite{Kveton2013} should be used to capture  the independent relationships. When the conditional independencies between the state variables are unknown in advance, kernel learning techniques need to be employed. By defining a space of kernel structures which are built compositionally from a context free grammar, we proposed a greedy search algorithm based on the previous works~\cite{Grosse2012,Duvenaud2013} to search over the grammar and automatically choose the decomposition structure from raw data by evaluation only a small fraction of all structures. We plan to demonstrate how the learned structure can represent and approximate the original RL problem in terms of compactness and efficiency, and evaluate our method on a
synthetic problem and compare it to other RL baselines.
\newpage
\section{Related Work and Future Challenges}
The representation learning methods described in this report can be applied to build representations from sampled examples over a large variety of problems in AI. They are also close related to recent work on manifold learning~\cite{Roweis2000,Belkin2004} and spectral learning~\cite{Narayanan2006}, which have largely been applied to nonlinear dimensionality reduction and semi-supervised learning problems on graphs. However, learning the compact MDP representation introduces new challenges not represented in supervised learning and  dimensionality reduction, as the set of training examples is not available as a batch, but must be collected through active exploration of the state space. Another challenge for representation learning in reinforcement learning is how well a compact representation transfers from one problem to another.

\bibliographystyle{acm}
\bibliography{general}

\end{document}